\documentclass[10pt,twocolumn,letterpaper]{article}

\usepackage{cvpr}
\usepackage{times}
\usepackage{epsfig}
\usepackage{graphicx}
\usepackage{amsmath}
\usepackage{amssymb}

\usepackage{comment}
\usepackage{url}

\usepackage{multirow}

\usepackage[pagebackref=true,breaklinks=true,letterpaper=true,colorlinks,bookmarks=false]{hyperref}

\cvprfinalcopy 

\def\cvprPaperID{****} 

\ifcvprfinal\fi
\begin{document}

\title{O-HAZE: a dehazing benchmark with real hazy and haze-free outdoor images}

\author{Codruta O. Ancuti$^{*}$, Cosmin Ancuti$^{*}$, Radu Timofte$^{\dag}$ and Christophe De Vleeschouwer$^{\ddag}$\\
$^{*}$MEO, Universitatea Politehnica Timisoara, Romania\\
$^{\dag}$ETH Zurich, Switzerland and Merantix GmbH, Germany\\
$^{\ddag}$ICTEAM, Universite Catholique de Louvain, Belgium
}

\maketitle

\begin{abstract}
Haze removal or dehazing  is a challenging ill-posed problem that has drawn a significant attention in the last few years. Despite this growing interest, the scientific community is still lacking a reference dataset to evaluate objectively and quantitatively the performance of proposed dehazing methods. The few datasets that are currently considered, both for assessment and  training of learning-based dehazing techniques, exclusively rely on synthetic hazy images. To address this limitation, we introduce the first outdoor scenes database (named O-HAZE) composed of pairs of real hazy and corresponding haze-free images. In practice, hazy images have been captured in presence of real haze, generated by professional haze machines, and O-HAZE contains  45 different  outdoor scenes depicting the same visual content recorded in haze-free and hazy conditions, under the same illumination parameters. To illustrate its usefulness, O-HAZE is used to compare a representative set of state-of-the-art dehazing techniques, using traditional image quality metrics such as PSNR, SSIM and CIEDE2000. This reveals the limitations of current techniques, and questions some of their underlying assumptions.
\end{abstract}


\section{Introduction}

Haze is a common atmospheric phenomena produced by small floating particles (aerosols) that reduce the visibility  of distant objects due to light scattering and attenuation. This results in a loss of local contrast for distant objects, in the addition of noise to the image, and in a selective attenuation of the light spectrum. The dehazing problem has been largely investigated in previous works.  Earlier approaches rely on atmospheric cues~\cite{Cozman_Krotkov_97,Narasimhan_2002}, on multiple images captured with polarization filters~\cite{PAMI_2003_Narasimhan_Nayar,Schechner_2003}, or on depth knowledge prior~\cite{Kopf_DeepPhoto_SggAsia2008,Tarel_ICCV_2009}. 
In contrast, single image dehazing, meaning dehazing without side information related to the scene geometry or to the atmospheric conditions, is a problem that has only been investigated in the past ten years.  Single image dehazing is mathematically ill-posed, because the degradation caused by haze is different for every pixel, and depends on the distance between the scene point and the camera. This dependency is generally expressed by the simplified but realistic light propagation model of Koschmieder~\cite{Koschmieder_1924}, combining transmission and airlight to describe how haze impacts the observed image.
According to this model, due to the atmospheric particles that absorb and scatter light, only a fraction of the reflected light reaches the observer. The light intensity $\mathcal{I}(x)$  reaching a pixel coordinate $x$ after passing a hazy medium is expressed as:
\begin{equation}
	\mathcal{I}(x)=\mathcal{J}(x) \enspace T \left(x \right)+A_\infty \enspace \left[ 1-T\left(x\right) \right]
\label{model}
\end{equation} 
where the haze-free reflectance is denoted by $\mathcal{J}(x)$, while $T(x)$ denotes the \textit{transmission} (in a uniform medium, it decreases exponentially with the depth) and  $A_\infty$ corresponds to the atmospheric light (a color constant).

Single image dehazing directly builds on this optical model. Recently many strategies~\cite{Fattal_Dehazing,Tan_Dehazing,Dehaze_He_CVPR_2009,Tarel_ICCV_2009,Kratz_and_Nishino_2009,Dehaze_Ancuti_ACCV,Codruta_ICIP_2010,Ancuti_TIP_2013,Fattal_Dehazing_TOG2014,Ancuti_GRSL_2014,Emberton_2015,Tang_2014} addressed this problem , by considering different kinds of priors to estimate the transmission.

Despite this prolific work, both the validation and the comparison of dehazing methods remain largely unsatisfactory, due to the absence of pairs of corresponding hazy and haze-free ground-truth images. The absence of the reference (haze-free) images in real-life scenarios justifies why most of the existing evaluation methods are based on non-reference image quality assessment (NR-IQA) strategies, leading to a lack of consensus about dehazing methods quality.

Today, all assessment datasets\cite{Tarel_2012,D_Hazy_2016,HazeRD_2017} rely on synthesized hazy images using a simplified optical model, and known depth. The use of synthetic data is due to the practical issues associated to the recording of reference and hazy images under identical illumination condition.  

\begin{figure*}[t!]
  \centering
  \includegraphics[width=1\linewidth]{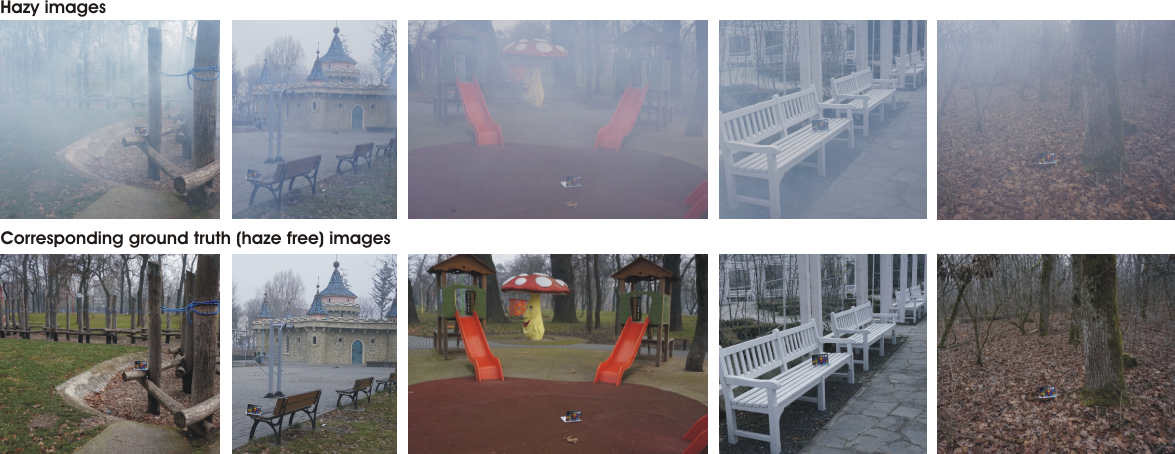}
  \caption{\label{fig:intro}%
     \textit{O-HAZE dataset provides 45 pairs of hazy and corresponding haze-free, i.e. groundtruth, outdoor images.} 
  }  
\end{figure*}

As an alternative, in this paper, we introduce O-HAZE, a dataset containing pairs of real hazy and corresponding haze-free images for $45$ various outdoor scenes. 
Haze has been generated with a professional haze machine that imitates with high fidelity real hazy conditions. 
Another contribution of this paper is a comprehensive evaluation of several state-of-the-art single image dehazing methods. 
Interestingly, our work reveals that many of the existing dehazing techniques are not able to accurately reconstruct the original image from its hazy version. This observation is founded on SSIM~\cite{Wang_2006} and CIEDE2000~\cite{Sharma_2005} objective image quality metrics, computed using the known reference and the dehazed results produced by different dehazing techniques. This observation, combined with the release of our dataset, should certainly motivate the design of improved dehazing methods.

The remainder of the paper is structured as follows. Section 2 surveys the previous work related to dehazing techniques and datasets. Section 3 describes the steps followed to record the O-HAZE dataset. Section 4 then briefly introduces the dehazing techniques that are quantitatively compared based on O-HAZE in Section 5.

\section{Related Work}

\subsection{Image dehazing techniques}

Image dehazing has drawn much attention in the last decade. 

To circumvent the ill-posed nature of the dehazing problem, early solutions~\cite{PAMI_2003_Narasimhan_Nayar} have built on multiple images obtained in different atmospheric conditions. Alternatively, Cozman and Krotkov~\cite{Cozman_Krotkov_97} and Nayar and Narasimhan~\cite{Narasimhan_2002} recover the visibility of hazy scenes by employing side atmospheric cues that facilitate image depth estimation. Other approaches rely on the registration of geometric models to get access to image depth~\cite{Kopf_DeepPhoto_SggAsia2008}. Hardware-based solutions have also been considered, such as the ones based on polarization filters~\cite{Chenault_2000,Schechner_2003,Schechner_2005}, which capture multiple images of the same scene with different degrees of polarization (DOP) and compare them to estimate the haze properties.

More recently, single image dehazing~\cite{Fattal_Dehazing,Tan_Dehazing,Dehaze_He_CVPR_2009,Tarel_ICCV_2009,Kratz_and_Nishino_2009,Dehaze_Ancuti_ACCV,Codruta_ICIP_2010,Ancuti_TIP_2013,Fattal_Dehazing_TOG2014,Ancuti_GRSL_2014,Emberton_2015,Tang_2014}  involves extensive research attention since it does not require special setup or side information, and is thus suitable even for challenging situations such as dynamic scenes. 
The pioneer work of Fattal~\cite{Fattal_Dehazing} accounts for surface shading in addition to the transmission function, and searches for a solution in which the resulting shading and transmission functions are locally statistically uncorrelated. Tan's approach~\cite{Tan_Dehazing} considers the brightest value in the image to estimate the atmospheric airlight, and directly searches to maximize the contrast of the output based on the assumptions that a haze-free image presents higher contrast compared with its hazy version. Contrast maximization has also been employed in~\cite{Tarel_ICCV_2009}, but with a lower computational complexity. In~\cite{Dehaze_Ancuti_ACCV}, a per-pixel identification of hazy regions is presented, using hue channel analysis and a \textit{semi-inverse} of the image.
The dark channel priors, introduced in~\cite{Dehaze_He_CVPR_2009},  has been proven to be a really effective solution to estimate the transmission, and has motivated many recent approaches. Meng et al.~\cite{Meng_2013} extends the work on dark channel, by regularizing the transmission around boundaries to mitigate its lack of resolution. Zhu et al.~\cite{Zhu_2015} extend~\cite{Meng_2013} by considering a color attenuation prior, assuming that the depth can be estimated from pixel saturation and intensity. Additionally, a standard median filter has been considered in Gibson et al.~\cite{Gibson_2012} to avoid halo artifacts in the dehazed image. The color-lines model~\cite{Fattal_Dehazing_TOG2014} employs a Markov Random Field (MRF) to  generate complete and regularized transmission maps. Similarly, the haze influence over the color channels distribution has also been exploited in~\cite{Berman_2016}, by observing that the colors of a haze-free image are well approximated by a limited set of tight clusters in the RGB space. In presence of haze, those clusters are spread along lines in the RGB space, as a function of the distance map, which allows to recover the haze free image. For night-time dehazing and non-uniform lighting conditions, spatially varying airlight has been considered and estimated in~\cite{Berman_2016, Ancuti_NightTime, Li_2015}.
Fusion-based single image dehazing approaches~\cite{Ancuti_TIP_2013,Choi_2015} have also achieved visually pleasant results without explicit transmission estimation and, more recently, several machine learning based methods have been introduced~\cite{Tang_2014,Dehazenet_2016,Ren_2016}. DehazeNet~\cite{Dehazenet_2016} takes a hazy image as input, and outputs its medium transmission map that is subsequently used to recover a haze-free image via atmospheric scattering model. For its training, DehazeNet resorts to data that are synthesized based on the physical haze formation model. Ren et al.~\cite{Ren_2016} proposed a coarse-to-fine network consisting of a cascade of CNN layers, also trained with synthesized hazy images.

\subsection{Dehazing assessment}
Despite the impressive effort in designing dehazing algorithms, a major issue preventing further developments is related to the impossibility to reliably asses the dehazing performance of a given algorithm, due to the absence of reference haze-free images (ground-truth). A key problem in collecting pairs of hazy and haze-free ground-truth images lies in the need to capture both images with identical scene illumination.

Due to this limitation, most dehazing quality metrics are restricted to non-reference image quality metrics (NR-IQA)~\cite{Mittal_2012,Mittal_2013,Saad_2012}. For example, in~\cite{Hautiere_2008}, the assessment simply relies on the gradient of the visible edges. A more general framework has been introduced in~\cite{Chen_2014}, using subjective assessment of enhanced and original images captured in bad visibility conditions. Besides, the Fog Aware Density Evaluator (FADE) introduced in~\cite{Choi_2015} predicts the visibility of a hazy/foggy scene from a single image without corresponding ground-truth. Unfortunately, due to the absence of the reference (haze-free) images in real-life scenarios, none of these approaches has been generally accepted by the dehazing community..

Another assessment strategy builds on synthesized hazy images, using the optical model and known depth to synthesize the haze effect.  
The work~\cite{Tarel_2012} presents the FRIDA dataset designed for Advanced Driver Assistance Systems (ADAS) that is a synthetic image database with 66 computer graphics generated roads scenes. In~\cite{D_Hazy_2016}, a dataset of 1400+ images of real complex scenes has been derived from the \textit{Middleburry}\footnote{\scriptsize{\url{http://vision.middlebury.edu/stereo/data/scenes2014/}}}  and the \textit{NYU-Depth V2}\footnote{\scriptsize{\url{http://cs.nyu.edu/~silberman/datasets/nyu_depth_v2.html}}} datasets. It contains high quality real scenes, and the depth map associated to each image has been used to yield synthesized hazy images based on Koschmieder's light propagation model~\cite{Koschmieder_1924}. This dataset has been recently extended in~\cite{HazeRD_2017}, by adding several synthesized outdoor hazy images. Interestingly, the work in J.E. Khoury et al.~\cite{Khoury_2016} introduces the CHIC (Color Hazy Image for Comparison) database, providing hazy and haze-free images in real scenes captured under controlled illumination. The dataset however only considers two indoor scenes, thereby failing to cover a large variation of textures and scene depth. To complete this preliminary achievement, our paper introduce a novel dataset that can be employed as a more representative benchmark to assess dehazing algorithms in outdoor scenes, based on ground truth images. The O-HAZE dataset includes $45$ various outdoor scenes, captured under controlled illumination. It thereby provides relevant data to support dehazing assessment, and is used in the NTIRE dehazing challenge\footnote{\scriptsize{\url{www.vision.ee.ethz.ch/ntire18/}}}.

\section{O:HAZE: Recording the outdoor hazy scenes}

The O-HAZE database has been derived from 45 various outdoor scenes in presence or absence of haze.  Our dataset allows to investigate the contribution of the haze over the scene visibility by analyzing  the scene objects radiance starting from the camera proximity to a maximum distance of 30m. 

Since we aimed for outdoor conditions similar to the ones encountered in hazy days, the recording period has been spread over more than 8 weeks during the autumn season.  
We have recorded the scenes during cloudy days,  in the morning or in the sunset, and only when the wind speed was below 3 km/h (to limit fast spreading of the haze in the scene). The absence of wind was the parameter that was the harder to meet, and explain why the recording of 45 scenes took more than 8 weeks.  In terms of hardware, we used a setup composed of a tripod and a Sony A5000 camera that was remotely controlled (Sony RM-VPR1). We acquired JPG and ARW (RAW) $5456 \times 3632$ images, with 24 bit depth. 

Each scene acquisition has started with a manual adjustment of the camera settings. The same parameters are adopted to capture the haze-free and hazy scene.  Those parameters include the shutter-speed (exposure-time), the aperture (F-stop), the ISO and white-balance. The similarity between hazy and haze-free acquisition settings is confirmed by the fact that the closer regions (that in general are less distorted by haze) have similar appearance (in terms of color and visibility) in the pair of hazy and haze-free images associated to a given scene.    

To set the camera parameters (aperture-exposure-ISO), we took advantage of the built-in light-meter of the camera, but also used an external exponometer (Sekonic).  For the custom white-balance, we used the middle gray card of the color checker. This commonly used  process  in photography is quite straight-forward and requires to change the camera white-balance mode in manual mode and place the reference grey-card in the front of it. For this step, we have placed the gray-card in the center of the scene in the range of four meters.  

After carefully checking that all the conditions above mentioned and after the setting procedure we have placed in each scene a color checker (Macbeth color checker) to allow for  post-processing of the recorded images. We use a classical Macbeth color checker with the size 11 by 8.25 inches with 24 squares of painted samples (4$\times$6 grid).
  
The haze was introduced in the scenes using two professional haze  machines (LSM1500 PRO 1500 W), which generate vapor particles with diameter size (typically 1 - 10 microns) similar to the particles of the atmospheric haze.  The haze  machines use cast or platen type aluminum heat exchangers to induce liquid evaporation. We have chosen special (haze) liquid with higher density in order to simulate the effect occurring with water haze over larger distances than the investigated 20-30 meters. The generation of haze took approximately 3-5 minutes. After haze generation, we used a fan to spread the haze as uniformly as possible in the scene.

\section{Evaluated Dehazing Techniques}

In this section, we briefly survey the state-of-the-art single image dehazing techniques that are compared based on the O-HAZE dataset. 
\newline
The first approach is the seminal work of \textbf{He et al.}~\cite{Dehaze_He_CVPR_2009}. This method introduces the dark channel prior (DCP),  a statistic observation  employed  by many recent dehazing techniques to yield a first estimate of the transmission map. DCP has been influenced by the dark object~\cite{Chavez_1988} and uses the key-observation that most of the local regions (with the exception of the sky or hazy regions) contain pixels that present low intensity in at least one of the color channels. In order to reduce artifacts, the transmission estimate is further refined based on alpha matting strategy~\cite{Dehaze_He_CVPR_2009} or guided filter~\cite{Guided_filter_PAMI_2013}. In our study, the results have been generated using the dark channel prior refined with the guided filter approach.\\   
\newline
\textbf{Meng et al.}~\cite{Meng_2013} further exploits the advantages of the dark channel prior~\cite{Dehaze_He_CVPR_2009}. The method applies a boundary constraint to the  transmission estimate yielded by DCP. The boundary constraint is  combined with a weighted L1−norm regularization. Overall, it mitigates the lack of resolution in teh DCP transmission map. This algorithm shows some improvement compared with the He et al.~\cite{Dehaze_He_CVPR_2009} technique reducing the level of the  halo artifacts around sharp edges. It also better  handles the appearance of the bright sky region.\\
\newline
\textbf{Fattal}~\cite{Fattal_Dehazing_TOG2014} introduces a method relying on the observation that the distributions of pixels in small natural image patches exhibit one-dimensional structures, named color-lines, in the RGB color space~\cite{Omer_2004}. A first  transmission estimation is computed from the detected color-lines offset to the origin, while a refined transmission is generated by a Markov random field model in charge of filtering the noise and removing other artifacts caused by scattering. \\
\newline
\textbf{Cai et al.}~\cite{Dehazenet_2016}  propose to adopt an end-to-end CNN deep model, trained to map hazy to haze-free patches. The algorithm is divided into four sequential steps: features extraction, multi-scale mapping, local extrema and finally non-linear regression. The training is based on synthesized hazy images. \\
\newline
\textbf{Ancuti et al.}~\cite{Ancuti_NightTime} introduce a novel straightforward method for local airlight estimation, and take  advantage of the  multi-scale fusion strategy to fusion the multiple versions obtained from distinct definitions of the locality notion. Although the solution has been developed to solve the complex night-time dehazing challenge, including in presence of severe scattering or multiple sources of light, the approach is also suited to day-time hazy scene enhancement. \\ 
\newline
\textbf{Berman et al.}~\cite{Berman_2016} further exploits the color consistency observation of~\cite{Omer_2004}, which considers that the color distribution in a haze-free images is well approximated by a discrete set of clusters in the RGB  colorspace. They observe that, in general, the pixels in a given cluster are non-local and are spread over the entire image plane. The pixels in a cluster are thus affected differently by the haze. As a consequence, each cluster becomes a line in the hazy image, and the position of a pixel within the line reflects its transmission level. In other words, these haze-lines convey information about the transmission in different regions of the image, and are used to estimate the transmission map.\\ 
\newline
\textbf{Ren et al.}~\cite{Ren_2016} develop a multi-scale CNN to estimate the transmission map directly from hazy images. The transmission map is first computed by a coarse-scale network, and is then refined by a fine-scale network. The training has been performed using synthetically generated hazy images, obtained from haze-free images and using their associated depth maps to apply a simplified light propagation model. 
\newline

\begin{figure*}[t!]
  \centering
  \includegraphics[width=1\linewidth]{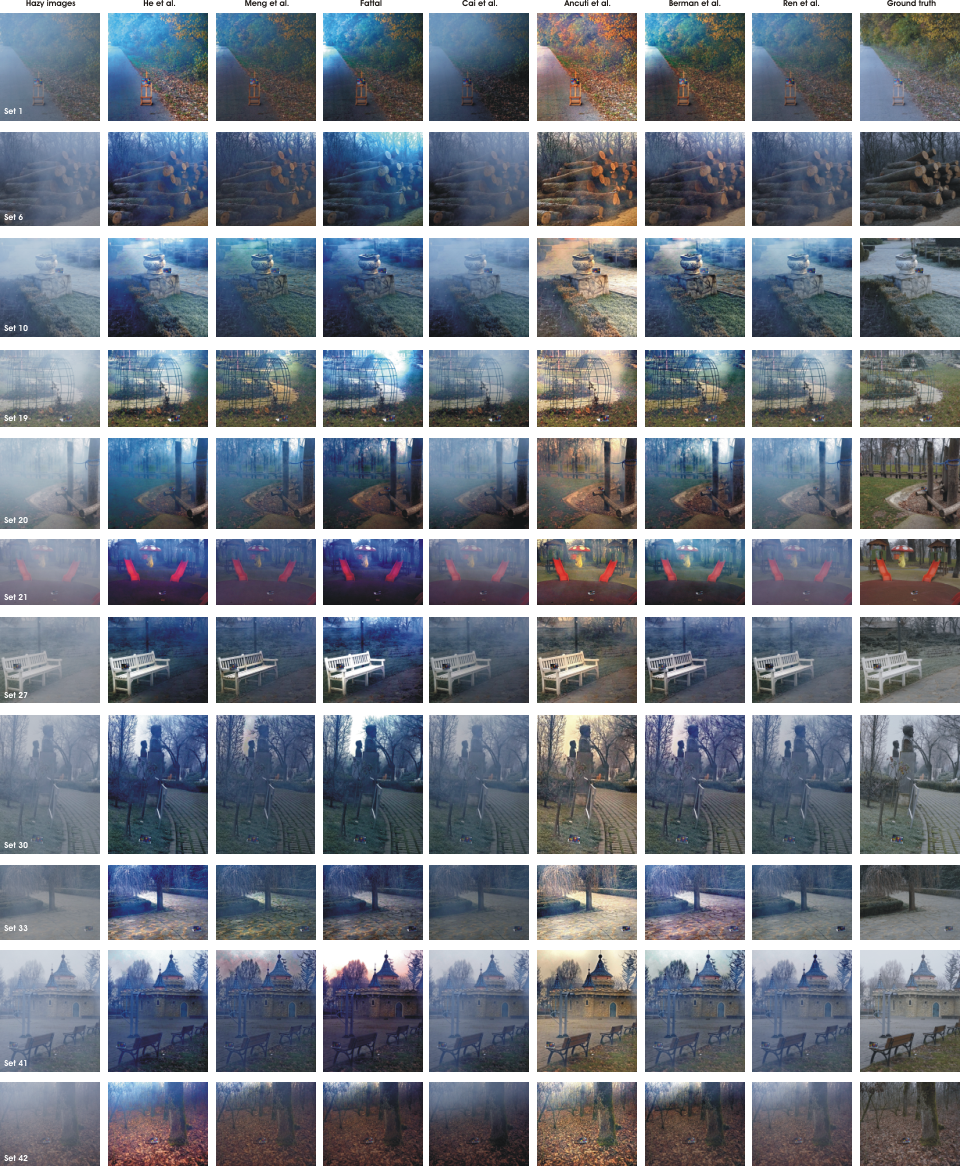}
  \caption{\label{fig:res_comp1}%
    \textit{\textbf{Comparative results.} The first row shows the hazy images and the last row shows  the ground truth. The other rows from left to right show the results of He et al.~\cite{Dehaze_He_CVPR_2009}, Meng et al.~\cite{Meng_2013}, Fattal~\cite{Fattal_Dehazing_TOG2014}, Cai et al.~\cite{Dehazenet_2016}, Ancuti et al.~\cite{Ancuti_NightTime},  Berman et al.~\cite{Berman_2016} and Ren et al.~\cite{Ren_2016}.}
  }  
\end{figure*}

\begin{table*}[]
\centering
\begin{tabular}{|l|l|l|l|l|l|l|l|l|l|l|l|l|l|l|}
\hline
\multirow{2}{*}{} & \multicolumn{2}{l|}{\small{\textbf{He et al.}}~\cite{Dehaze_He_CVPR_2009}} & \multicolumn{2}{l|}{\small{\textbf{Meng et al.}}~\cite{Meng_2013}} & \multicolumn{2}{l|}{\small{\textbf{Fattal}}~\cite{Fattal_Dehazing_TOG2014}} & \multicolumn{2}{l|}{\small{\textbf{Cai et al.}}~\cite{Dehazenet_2016}} & \multicolumn{2}{l|}{\small{\textbf{Ancuti et al.}}~\cite{Ancuti_NightTime}} & \multicolumn{2}{l|}{\small{\textbf{Berman et al.}}~\cite{Berman_2016}} & \multicolumn{2}{l|}{\small{\textbf{Ren et al.}}~\cite{Ren_2016}} \\ \cline{2-15} 
                  & {\scriptsize{SSIM}}             & {\tiny{CIEDE2000}}            & {\scriptsize{SSIM}}              & {\tiny{CIEDE2000}}            & {\scriptsize{SSIM}}            & {\tiny{CIEDE2000}}          & {\scriptsize{SSIM}}              & {\tiny{CIEDE2000}}            & {\scriptsize{SSIM}}               & {\tiny{CIEDE2000}}              & {\scriptsize{SSIM}}               & {\tiny{CIEDE2000}}              & {\scriptsize{SSIM}}              & {\tiny{CIEDE2000}}            \\ \hline
\textbf{Set 1}    & 0.82             & 22.37                & 0.77                  & 21.06            & 0.73           & 24.29               & 0.58             & 24.42                 & 0.75                 & 20.09                & 0.76                 & 20.97                & \textbf{0.81}      & \textbf{18.17}      \\ \hline
\textbf{Set 6}    & 0.74             & 19.00                & \textbf{0.78}         & 11.44            & 0.73           & 21.89               & 0.59             & 16.16                 & 0.68                 & 15.53                & 0.77                 & \textbf{12.68}       & 0.72               & 13.20               \\ \hline
\textbf{Set 10}   & 0.78             & 15.22                & 0.76                  & 16.63            & 0.75           & 17.49               & 0.71             & 16.17                 & 0.73                 & 19.21                & 0.72                 & 17.77                & \textbf{0.80}      & \textbf{13.70}      \\ \hline
\textbf{Set 19}   & 0.81             & 16.31                & \textbf{0.84}         & 13.37            & 0.79           & 21.48               & 0.72             & 16.92                 & 0.78                 & 15.55                & 0.82                 & 14.49                & 0.83               & \textbf{12.94}      \\ \hline
\textbf{Set 20}   & 0.61             & 23.81                & 0.72                  & 20.91            & 0.62           & 20.73               & 0.50             & 23.71                 & \textbf{0.78}        & \textbf{12.67}       & 0.72                 & 19.40                & 0.63               & 20.98               \\ \hline
\textbf{Set 21}   & 0.69             & 27.50                & 0.78                  & 21.13            & 0.63           & 28.25               & 0.71             & 19.49                 & \textbf{0.78}        & \textbf{10.72}       & 0.72                 & 20.54                & 0.73               & 20.26               \\ \hline
\textbf{Set 27}   & 0.61             & 21.38                & 0.68                  & 18.76            & 0.67           & 22.37               & 0.64             & 17.16                 & \textbf{0.77}        & \textbf{10.94}       & 0.70                 & 18.41                & 0.71               & 14.16               \\ \hline
\textbf{Set 30}   & 0.75             & 18.85                & 0.74                  & 18.59            & 0.72           & 18.46               & 0.77             & 12.70                 & \textbf{0.83}        & \textbf{11.25}       & 0.81                 & 14.55                & 0.82               & 12.66               \\ \hline
\textbf{Set 33}   & 0.76             & 18.54                & 0.74                  & 15.84            & 0.76           & 17.86               & 0.81             & 14.61                 & 0.61                 & 20.86                & 0.66                 & 19.39                & \textbf{0.88}      & \textbf{10.87}      \\ \hline
\textbf{Set 41}   & 0.77             & 19.54                & 0.72                  & 21.45            & 0.66           & 23.71               & 0.84             & 12.78                 & 0.84                 & 13.02                & 0.82                 & 14.36                & \textbf{0.88}      & \textbf{12.34}      \\ \hline
\textbf{Set 42}   & 0.79             & 19.70                & 0.82                  & 11.03            & 0.73           & 13.21               & 0.58             & 15.58                 & 0.74                 & 15.37                & \textbf{0.82}        & \textbf{11.00}       & 0.72               & 12.87               \\ \hline
\end{tabular}
\caption{\label{tabel_eval1} \textit{\textbf{Quantitative evaluation.} We randomly picked up 11 sets  from our O-HAZE dataset, and did compute the  SSIM and CIEDE2000 indices between the ground truth images and the dehazed images produced by the evaluated techniques. The hazy images, ground truth and the results are shown in Fig.\ref{fig:res_comp1}.}}
\end{table*}

\begin{table*}[]
\centering
\label{table_average}
\begin{tabular}{|l|l|l|l|l|l|l|l|}
\hline
          & {\small{\textbf{He et al.}}~\cite{Dehaze_He_CVPR_2009}} & {\small{\textbf{Meng et al.}}~\cite{Meng_2013}} & {\small{\textbf{Fattal}}~\cite{Fattal_Dehazing_TOG2014}} & {\small{\textbf{Cai et al.}}~\cite{Dehazenet_2016}} & {\small{\textbf{Ancuti et al.}}~\cite{Ancuti_NightTime}} & {\small{\textbf{Berman et al.}}~\cite{Berman_2016}} & {\small{\textbf{Ren et al.}}~\cite{Ren_2016}} \\ \hline
\small{\textbf{SSIM}}      & 0.735     & 0.753      & 0.707  & 0.666      & 0.747         & 0.750         & 0.765      \\ \hline
\small{\textbf{PSNR}}      & 16.586    & 17.443     & 15.639 & 16.207     & 16.855        & 16.610        & 19.068     \\ \hline
\small{\textbf{CIEDE2000}} & 20.745    & 16.968     & 19.854 & 17.348     & 16.431        & 17.088        & 14.670     \\ \hline
\end{tabular}
\caption{Quantitative evaluation of all the 45 set of images of the O-HAZE dataset. This table presents the average values of the SSIM, PSNR and CIEDE2000 indexes, over the entire dataset. }
\end{table*}

\section{Evaluation and Discussion}

\begin{figure*}[t!]
  \centering
  \includegraphics[width=1\linewidth]{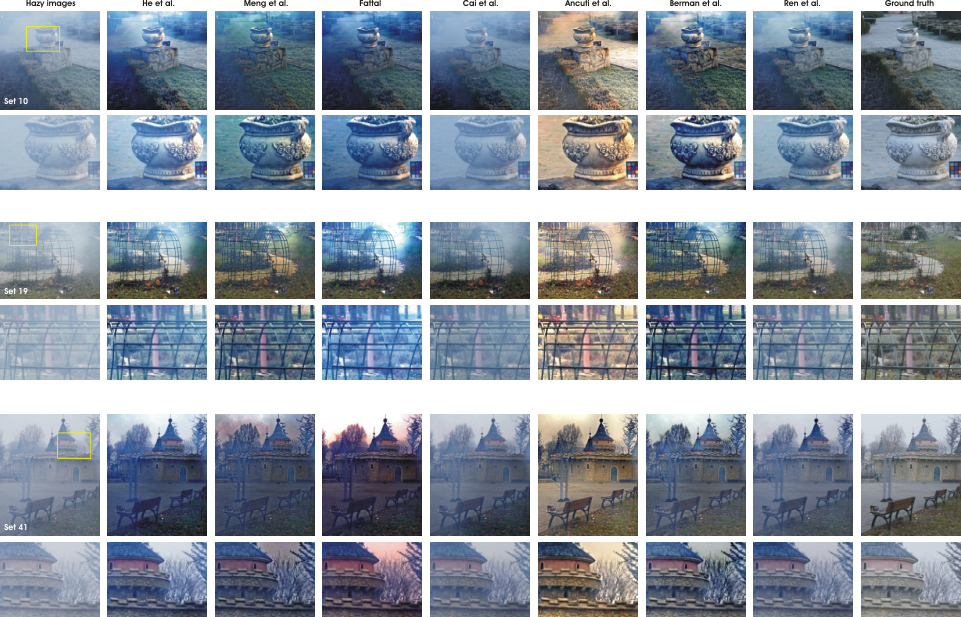}
  \caption{\label{fig:res_comp_crop1}%
    \textit{\textbf{Comparative detail insets.} The first, third and fifth rows show  the hazy images (first column), their corresponding ground truth (last column), and the results of several dehazing techniques~\cite{Dehaze_He_CVPR_2009,Meng_2013,Fattal_Dehazing_TOG2014,Dehazenet_2016,Ancuti_NightTime,Berman_2016,Ren_2016}, for the scenes 10, 19, and 41, respectively. The corresponding detail insets are shown below in the even rows. }
  }  
\end{figure*}

We have used our proposed dataset to perform a comprehensive evaluation of the state-of-the-art single image dehazing techniques presented in Section 4. Fig.~\ref{fig:res_comp1} depicts 11 scenes, randomly extracted from the O-HAZE dataset. Each hazy image is shown in the first column, and  the corresponding ground truth (haze-free) in the last column.  The other columns, from left to right, present the results yielded by the techniques of He et al.~\cite{Dehaze_He_CVPR_2009}, Meng et al.~\cite{Meng_2013}, Fattal~\cite{Fattal_Dehazing_TOG2014}, Cai et al.~\cite{Dehazenet_2016}, Ancuti et al.~\cite{Ancuti_NightTime},  Berman et al.~\cite{Berman_2016} and Ren et al.~\cite{Ren_2016}. Additionally, Fig. \ref{fig:res_comp_crop1} presents the comparative detail insets of scenes 10, 19 and 41, respectively.

Qualitatively, we can observe that the results of He et al.~\cite{Dehaze_He_CVPR_2009} recover quite well the image structure, but introduce unpleasing color shifting in the hazy regions mostly due to the poor airlight estimation. As expected, these distortions are more significant in the lighter/whiter regions, where the dark channel prior in general fails. 
The method  proposed by Meng et al.~\cite{Meng_2013} also rely on the dark channel prior. It improves the results obtained by He et al.~\cite{Dehaze_He_CVPR_2009} thanks to a more accurate transmission estimation.  The color lines exploited in Fattal~\cite{Fattal_Dehazing_TOG2014} also introduce some unpleasing color artifacts. The  method of Berman et al~\cite{Berman_2016} is less prone to such artifacts, and lead to images with sharp edges, mostly due to their strategy to locally estimate the airlight and the transmission. Due to the multi-scale fusion and local airlight estimation, the method of Ancuti et al.~\cite{Ancuti_NightTime} handles color differently than other methods, leading to higher contrast and more intense colors, but with a slight shift towards yellow/red. 

Regarding the learning-based approaches, we observe that the method of Ren et al~\cite{Ren_2016} generates visually more compelling results than the deep learning approach of Cai et al~\cite{Dehazenet_2016}. 

As a general conclusion, it appears that there is not a clear winner between the learning-based techniques~\cite{Dehazenet_2016,Ren_2016} and the algorithms that rely on explicit priors~\cite{Dehaze_He_CVPR_2009,Meng_2013,Fattal_Dehazing_TOG2014,Ancuti_NightTime,Berman_2016}. All the tested methods introduce structural distortions such as halo artifacts close to the edges, especially in regions that are far away from the camera. Moreover, color distortion occasionnally creates some unnatural appearances of the dehazed images. Hence, the algorithms of both classes introduce important structure artifacts and color shifting. More interestingly, despite none algorithm is able to recover the ground-truth image, we observe that some of the existing algorithms are complementary in the sense that they offer distinct and specific benefits, either in terms of color contrast or image sharpness, while suffering from different artifacts (see for example Ancuti et al.~\cite{Ancuti_NightTime} and Berman et al.~\cite{Berman_2016}). This certainly leaves room for improvement through appropriate combination of methods.

The O-HAZE dataset has the main advantage to allow for an objective quantitative evaluation of the dehazing techniques.  Table~\ref{tabel_eval1} compares the output of different dehazing techniques with the ground-truth (haze free) images based on PSNR, SSIM~\cite{Wang_2004} and CIEDE2000~\cite{Sharma_2005,Westland_2012}, for the images shown in  Fig.~\ref{fig:res_comp1}. The structure similarity index (SSIM) compares local patterns of pixel intensities that have been normalized for luminance and contrast. The SSIM ranges in [-1,1], with maximum value 1 for two identical images. In addition, to evaluate the level of color restoration, we compute the CIEDE2000~\cite{Sharma_2005,Westland_2012}.  CIEDE2000 measures accurately the color difference between two images and generates values in the range [0,100], with smaller values indicating better color preservation. 
In addition to Table~\ref{tabel_eval1}, Table~\ref{table_average} presents  the average SSIM, PSNR and CIEDE values over the entire 45 scenes of the O-HAZE dataset. From  these tables, we can conclude that in terms of structure and color restoration the methods of Meng et al.~\cite{Meng_2013}, Berman et al.~\cite{Berman_2016}, Ancuti et al.~\cite{Ancuti_NightTime}  and Ren et al.~\cite{Ren_2016} perform the best in average when considering the SSIM, PSNR and CIEDE indices. A second group of methods including He et al.~\cite{Dehaze_He_CVPR_2009}, Fattal~\cite{Fattal_Dehazing_TOG2014} and Cai et al~\cite{Dehazenet_2016}, are less competitive both in terms of structure and color restoration.\newline 
Overall, none of the techniques performs better than others for all images. The relatively poor SSIM, PSNR and CIEDE2000 quality metrics proove once again the difficulty of single image dehazing task, and the fact that there is still much room for improvement.\newline

{
\bibliographystyle{ieee}
\bibliography{ref}

\begin{thebibliography}{10}\itemsep=-1pt

\bibitem{Ancuti_TIP_2013}
C.~Ancuti and C.~Ancuti.
\newblock Single image dehazing by multi-scale fusion.
\newblock {\em IEEE Transactions on Image Processing}, 22(8):3271--3282, 2013.

\bibitem{Ancuti_GRSL_2014}
C.~Ancuti and C.~O. Ancuti.
\newblock Effective contrast-based dehazing for robust image matching.
\newblock {\em IEEE Geoscience and Remote Sensing Letters}, 2014.

\bibitem{Ancuti_NightTime}
C.~Ancuti, C.~O. Ancuti, A.~Bovik, and C.~D. Vleeschouwer.
\newblock Night time dehazing by fusion.
\newblock {\em IEEE ICIP}, 2016.

\bibitem{D_Hazy_2016}
C.~Ancuti, C.~O. Ancuti, and C.~D. Vleeschouwer.
\newblock D-hazy: A dataset to evaluate quantitatively dehazing algorithms.
\newblock {\em IEEE ICIP}, 2016.

\bibitem{Codruta_ICIP_2010}
C.~O. Ancuti, C.~Ancuti, and P.~Bekaert.
\newblock Effective single-image dehazing by fusion.
\newblock In {\em In IEEE ICIP}, 2010.

\bibitem{Dehaze_Ancuti_ACCV}
C.~O. Ancuti, C.~Ancuti, C.~Hermans, and P.~Bekaert.
\newblock A fast semi-inverse approach to detect and remove the haze from a
  single image.
\newblock {\em ACCV}, 2010.

\bibitem{Berman_2016}
D.~Berman, T.~Treibitz, and S.~Avidan.
\newblock Non-local image dehazing.
\newblock {\em IEEE Intl. Conf. Comp. Vision, and Pattern Recog}, 2016.

\bibitem{Dehazenet_2016}
B.~Cai, X.~Xu, K.~Jia, C.~Qing, and D.~Tao.
\newblock Dehazenet: An end-to-end system for single image haze removal.
\newblock {\em IEEE Transactions on Image Processing}, 2016.

\bibitem{Chavez_1988}
P.~Chavez.
\newblock An improved dark-object subtraction technique for atmospheric
  scattering correction of multispectral data.
\newblock {\em Remote Sensing of Environment}, 1988.

\bibitem{Chen_2014}
Z.~Chen, T.~Jiang, and Y.~Tian.
\newblock Quality assessment for comparing image enhancement algorithms.
\newblock {\em In IEEE Conference on Computer Vision and Pattern Recognition},
  2014.

\bibitem{Chenault_2000}
D.~Chenault and J.~Pezzaniti.
\newblock Polarization imaging through scattering media.
\newblock {\em In Proc. SPIE}, 2000.

\bibitem{Choi_2015}
L.~K. Choi, J.~You, and A.~C. Bovik.
\newblock Referenceless prediction of perceptual fog density and perceptual
  image defogging.
\newblock {\em In IEEE Trans. on Image Processing}, 2015.

\bibitem{Cozman_Krotkov_97}
F.~Cozman and E.~Krotkov.
\newblock Depth from scattering.
\newblock {\em IEEE Conf. Computer Vision and Pattern Recognition}, 1997.

\bibitem{Emberton_2015}
S.~Emberton, L.~Chittka, and A.~Cavallaro.
\newblock Hierarchical rank-based veiling light estimation for underwater
  dehazing.
\newblock {\em Proc. of British Machine Vision Conference (BMVC)}, 2015.

\bibitem{Fattal_Dehazing}
R.~Fattal.
\newblock Single image dehazing.
\newblock {\em SIGGRAPH}, 2008.

\bibitem{Fattal_Dehazing_TOG2014}
R.~Fattal.
\newblock Dehazing using color-lines.
\newblock {\em ACM Trans. on Graph.}, 2014.

\bibitem{Gibson_2012}
K.~B. Gibson, D.~T. Vo, and T.~Q. Nguyen.
\newblock An investigation of dehazing effects on image and video coding.
\newblock {\em IEEE Trans. Image Proc.}, 2012.

\bibitem{Hautiere_2008}
N.~Hautiere, J.-P. Tarel, D.~Aubert, and E.~Dumont.
\newblock Blind contrast enhancement assessment by gradient ratioing at visible
  edges.
\newblock {\em Journal of Image Analysis and Stereology}, 2008.

\bibitem{Dehaze_He_CVPR_2009}
K.~He, J.~Sun, and X.~Tang.
\newblock Single image haze removal using dark channel prior.
\newblock {\em In IEEE CVPR}, 2009.

\bibitem{Guided_filter_PAMI_2013}
K.~He, J.~Sun, and X.~Tang.
\newblock Guided image filtering.
\newblock {\em In IEEE Transactions on Pattern Analysis and Machine
  Intelligence (TPAMI)}, 2013.

\bibitem{Khoury_2016}
J.~E. Khoury(B), J.-B. Thomas, and A.~Mansouri.
\newblock A color image database for haze model and dehazing methods
  evaluation.
\newblock {\em ICISP}, 2016.

\bibitem{Kopf_DeepPhoto_SggAsia2008}
J.~Kopf, B.~Neubert, B.~Chen, M.~Cohen, D.~Cohen-Or, O.~Deussen,
  M.~Uyttendaele, and D.~Lischinski.
\newblock Deep photo: Model-based photograph enhancement and viewing.
\newblock In {\em Siggraph ASIA, ACM Trans. on Graph.}, 2008.

\bibitem{Koschmieder_1924}
H.~Koschmieder.
\newblock Theorie der horizontalen sichtweite.
\newblock In {\em Beitrage zur Physik der freien Atmosphare}, 1924.

\bibitem{Kratz_and_Nishino_2009}
L.~Kratz and K.~Nishino.
\newblock Factorizing scene albedo and depth from a single foggy image.
\newblock {\em ICCV}, 2009.

\bibitem{Li_2015}
Y.~Li, R.~T. Tan, and M.~S. Brown.
\newblock Nighttime haze removal with glow and multiple light colors.
\newblock {\em In IEEE Int. Conf. on Computer Vision}, 2015.

\bibitem{Meng_2013}
G.~Meng, Y.~Wang, J.~Duan, S.~Xiang, and C.~Pan.
\newblock Efficient image dehazing with boundary constraint and contextual
  regularization.
\newblock {\em In IEEE Int. Conf. on Computer Vision}, 2013.

\bibitem{Mittal_2012}
A.~Mittal, A.~K. Moorthy, and A.~C. Bovik.
\newblock No-reference image quality assessment in the spatial domain.
\newblock {\em In IEEE Trans. on Image Processing}, 2012.

\bibitem{Mittal_2013}
A.~Mittal, R.~Soundararajan, and A.~C. Bovik.
\newblock Making a completely blind image quality analyzer.
\newblock {\em In IEEE Signal Processing Letters}, 2013.

\bibitem{Narasimhan_2002}
S.~Narasimhan and S.~Nayar.
\newblock Vision and the atmosphere.
\newblock {\em Int. J. Computer Vision,}, 2002.

\bibitem{PAMI_2003_Narasimhan_Nayar}
S.~Narasimhan and S.~Nayar.
\newblock Contrast restoration of weather degraded images.
\newblock {\em IEEE Trans. on Pattern Analysis and Machine Intell.}, 2003.

\bibitem{Omer_2004}
I.~Omer and M.~Andwerman.
\newblock Color lines: image specific color representation.
\newblock {\em In IEEE Conference on Computer Vision and Pattern Recognition},
  2004.

\bibitem{Ren_2016}
W.~Ren, S.~Liu, H.~Zhang, X.~C. J.~Pan, and M.-H. Yang.
\newblock Single image dehazing via multi-scale convolutional neural networks.
\newblock {\em Proc. European Conf. Computer Vision}, 2016.

\bibitem{Saad_2012}
M.~A. Saad, A.~C. Bovik, and C.~Charrier.
\newblock Blind image quality assessment: A natural scene statistics approach
  in the dct domain.
\newblock {\em In IEEE Trans. on Image Processing}, 2012.

\bibitem{Schechner_2005}
Y.~Y. Schechner and N.~Karpel.
\newblock Recovery of underwater visibility and structure by polarization
  analysis.
\newblock {\em IEEE Journal of Oceanic Engineering}, 2005.

\bibitem{Schechner_2003}
Y.~Y. Schechner, S.~G. Narasimhan, and S.~K. Nayar.
\newblock Polarization-based vision through haze.
\newblock {\em Applied Optics}, 2003.

\bibitem{Sharma_2005}
G.~Sharma, W.~Wu, and E.~Dalal.
\newblock The ciede2000 color-difference formula: Implementation notes,
  supplementary test data, and mathematical observations.
\newblock {\em Color Research and Applications}, 2005.

\bibitem{Tan_Dehazing}
R.~T. Tan.
\newblock Visibility in bad weather from a single image.
\newblock {\em In IEEE Conference on Computer Vision and Pattern Recognition},
  2008.

\bibitem{Tang_2014}
K.~Tang, J.~Yang, and J.~Wang.
\newblock Investigating haze-relevant features in a learning framework for
  image dehazing.
\newblock {\em In IEEE Conference on Computer Vision and Pattern Recognition},
  2014.

\bibitem{Tarel_ICCV_2009}
J.-P. Tarel and N.~Hautiere.
\newblock Fast visibility restoration from a single color or gray level image.
\newblock {\em In IEEE ICCV}, 2009.

\bibitem{Tarel_2012}
J.-P. Tarel, N.~Hautière, L.~Caraffa, A.~Cord, H.~Halmaoui, and D.~Gruyer.
\newblock Vision enhancement in homogeneous and heterogeneous fog.
\newblock {\em IEEE Intelligent Transportation Systems Magazine}, 2012.

\bibitem{Wang_2006}
Z.~Wang and A.~C. Bovik.
\newblock Modern image quality assessment.
\newblock {\em Morgan and Claypool Publishers}, 2006.

\bibitem{Wang_2004}
Z.~Wang, A.~C. Bovik, H.~R. Sheikh, and E.~P. Simoncelli.
\newblock Image quality assessment: From error visibility to structural
  similarity.
\newblock {\em IEEE Transactions on Image Processing}, 2004.

\bibitem{Westland_2012}
S.~Westland, C.~Ripamonti, and V.~Cheung.
\newblock Computational colour science using matlab, 2nd edition.
\newblock {\em Wiley}, 2005.

\bibitem{HazeRD_2017}
Y.~Zhang, L.~Ding, and G.~Sharma.
\newblock Hazerd: an outdoor scene dataset and benchmark for single image
  dehazing.
\newblock {\em IEEE ICIP}, 2017.

\bibitem{Zhu_2015}
Q.~Zhu, J.~Mai, and L.~Shao.
\newblock A fast single image haze removal algorithm using color attenuation
  prior.
\newblock {\em IEEE Trans. Image Proc.}, 2015.

\end{thebibliography}
}

\end{document}